\documentclass[fleqn, 12pt]{wlscirep}
\usepackage{courier}
\usepackage[utf8]{inputenc}
\usepackage[T1]{fontenc}
\usepackage{amsmath, amssymb}
\usepackage{kotex}
\usepackage{setspace}
\usepackage{graphicx}
\usepackage{helvet}
\usepackage{hyperref}
\graphicspath{../figures}
\title{Bayesian neural network with pretrained protein embedding enhances prediction accuracy of drug-protein interaction}

\author[1]{QHwan Kim}
\author[1]{Joon-Hyuk Ko}
\author[1]{Sunghoon Kim}
\author[1]{Nojun Park}
\author[1*]{Wonho Jhe}
\affil[1]{Department of Physics and Astronomy, Institute of Applied Physics, Seoul National University, Gwanak-gu, Seoul 08826, Republic of Korea.}

\affil[*]{whjhe@snu.ac.kr}


\begin{abstract} 
The characterization of drug-protein interactions is crucial in the high-throughput screening for drug discovery. The deep learning-based approaches have attracted attention because they can predict drug-protein interactions without trial-and-error by humans. However, because data labeling requires significant resources, the available protein data size is relatively small, which consequently decreases model performance. Here we propose two methods to construct a deep learning framework that exhibits superior performance with a small labeled dataset. At first, we use transfer learning in encoding protein sequences with a pretrained model, which trains general sequence representations in an unsupervised manner. Second, we use a Bayesian neural network to make a robust model by estimating the data uncertainty. As a result, our model performs better than the previous baselines for predicting drug-protein interactions. We also show that the quantified uncertainty from the Bayesian inference is related to the confidence and can be used for screening DPI data points.

\end{abstract}

\begin{document}
	\sf

\flushbottom
\maketitle
%
%

 Identifying novel drug-protein interactions (DPIs) have been studied broadly for predicting potential side effects\cite{Mizutani2012}, toxicities\cite{Liebler2005}, and repositioning of drugs\cite{Pushpakom2018, Xue2018}. However, quantification of DPI of every possible drug-protein pair is prohibitively time-consuming and expensive since it requires individual experiments or simulations for each pair. 

 With the development of protein sequence and drug-protein interaction public datasets\cite{Liu2006, Liu2015}, machine learning-based methods\cite{Fokoue2016, He2017, Wen2017, Vamathevan2019} have emerged as candidate of fast DPI identification. Recently, deep neural networks (DNNs) have attracted attention because they outperform other machine learning-based methods in various tasks, such as computer vision\cite{He2015} and natural language processing\cite{Vaswani2017, Devlin2019}. 

 In the usual DPI task, a protein is represented as a 1-dimensional long sequence of amino acid characters. Thus the deep learning models for natural language processing have been broadly used to obtain useful protein features from the sequences. Previous studies in this approach include using recurrent neural networks with long short-term memory (LSTM)\cite{Hochreiter1997} or gated recurrent unit (GRU)\cite{Cho2014} layers for their ability to identify long-term dependencies in sequential data\cite{Gao2018, Karimi2019, Wang2020}. The other studies have used convolutional neural networks (CNN)\cite{Ozturk2018, Lee2019, Shin2019, Tsubaki2019, Zhang2019} to extract hidden local patterns in sequences. Different representations of pretiens, such as 2-dimensional contact maps\cite{Jiang2020, Zheng2020} or 3-dimensional atom coordinates\cite{Lim2019, Morrone2020}, in addition to 1-dimensional sequences, also have been used to increase model performance. 
  
Supervised training of high-capacity DNN models from scratch requires a large amount of labeled training data points. For example, Mahajan \textit{et al.}\cite{Mahajan2018} showed that the more labeled data is required to increase accuracy after training with $10^{9}$ images. However, currently available DPI datasets usually contain thousands of labeled protein sequences, a small number compared to the > 195 M unrevealed interaction information in UniProtKB\cite{Uniprot2015}. The lack of qualified labeled data points suppresses the usage of more elaborated deep learning architectures, which could potentially increase performance and reliability\cite{Brigato2020}. In particular, the scarcity of labeled data of biology and chemistry-related tasks has been suggested consistently\cite{Ryu2019, Vamathevan2019} although the labeling require expensive and time-consuming experiments.
  
 To overcome the difficulties of learning with limited data, several studies have proposed methods to increase the expressiveness of the deep learning model without additional endeavor to label generation. Of those, transfer learning uses a model that is initially pretrained with a large corpus of data of different task. This pretrained model is then transferred to the target tasks by adding classification layers and fine-tuning with the original small dataset. Transfer learning approaches have shown substantial performance improvement in computer vision\cite{Kornblith2019}, natural language processing\cite{Devlin2019}, and structure-property prediction of molecules\cite{Winter2019, Hu2020}. In cases where labeled data is expensive, such as in scientific problems, the pretrained model can be prepared in an unsupervised manner, using large but unlabeled datasets. Winter \text{et al.}\cite{Winter2019} trained an autoencoder model with a huge corpus of chemical structures and used them to predict molecular properties. Villegas-Morcillo \textit{et al.}\cite{Villegas-Morcillo2020} showed that the supervised classification tasks with a pretrained protein sequence model could achieve competitive performance with other complicated models. 
 
 Another proposed method to obtain a more robust and reliable model with a small dataset is Bayesian neural network (BNN)\cite{Gal2015}. Compared to a conventional DNN, which gives definite point prediction for each given input, a BNN returns a distribution of predictions, which qualitatively corresponds to the aggregate prediction of an ensemble of different neural networks trained on the same dataset. Direct implementation of BNN is infeasible because training an ensemble of neural networks requires enormous computing power. Monte-Carlo dropout (MC-dropout) approach\cite{Gal2016, Kendall2017} enables reasonable training time of BNNs by approximating the posterior distribution of network weights by a product of the Bernoulli distribution with dropout layers.
   
 Here, we propose an end-to-end deep learning framework for highly accurate DPI prediction with transfer learning and BNN. We choose the pretrained model as a stacked transformer architecture, which is trained with 250 million unlabelled protein sequences in an unsupervised manner\cite{Rives2019}. The drug is represented by the molecular graph and encoded through the graph interaction network layers. Estimation of the model performance using three public DPI datasets shows that the proposed model outperforms previous approaches. In addition, the estimated uncertainty, which is obtained from the sampling output of BNN, is decomposed into model-based and data-based elements, which can be used to further virtual screening of data points. In summary, the main contributions of our work are as follows.
\begin{enumerate}
\item We propose the first approach to predict DPI with the Bayesian neural network framework and the pretrained protein sequence model;
\item our method demonstrates highly accurate prediction of three public DPI datasets;
\item the output of BNN can estimate the confidence of the data points.
\end{enumerate}   

\section*{Experiments}

\subsection*{Datasets}
 We evaluate our model and other baseline models on three public DPI datasets: the BindingDB dataset\cite{Gao2018}, the Human dataset\cite{Liu2015}, and the \textit{C. elegans} dataset\cite{Liu2015}.

\subsubsection*{BindingDB}
 BindingDB is a public database of experimentally measured binding affinities between the small molecules and proteins\cite{Liu2006}. The original dataset contains 1.3 million interaction labels with quantitative measurements of $\textrm{IC}_{50}$, $\textrm{EC}50$, and $\textrm{Ki}$. We use the binarized version of BindingDB dataset constructed by Gao \textit{et al.}\cite{Gao2018}, which contains 39,747 positive interactions and 31,218 negative interactions. The training/validation/testing split is prepared in the dataset. The training set contains 28,240 positive and 21,915 negative interactions. The validation set contains 2,831 positive and 2,776 negative interactions. And the test set contains 2,706 positive and 2,802 negative interactions. 

\subsubsection*{Human and \textit{C. elegans}}
 Created by Liu \textit{et al.}\cite{Liu2015}, these datasets include highly credible negative samples of compound-protein pairs obtained by using a systematic screening framework. Following Tsubaki \textit{et al.}\cite{Tsubaki2019}, we use the balanced and the unbalanced dataset, where the ratios of the positive to negative samples are 1:1 and 1:3, respectively. The human dataset contains 3,369 positive interactions between 1,052 unique drugs and 852 unique proteins; the \textit{C. elegans} dataset contains 4,000 positive interactions between 1,434 unique drugs and 2,504 unique proteins. Also, we use an 80\%/10\%/10\% training/validation/testing random split.

 \subsection*{Proposed Model}  
\begin{figure}[h]
	\includegraphics[width=0.6\textwidth]{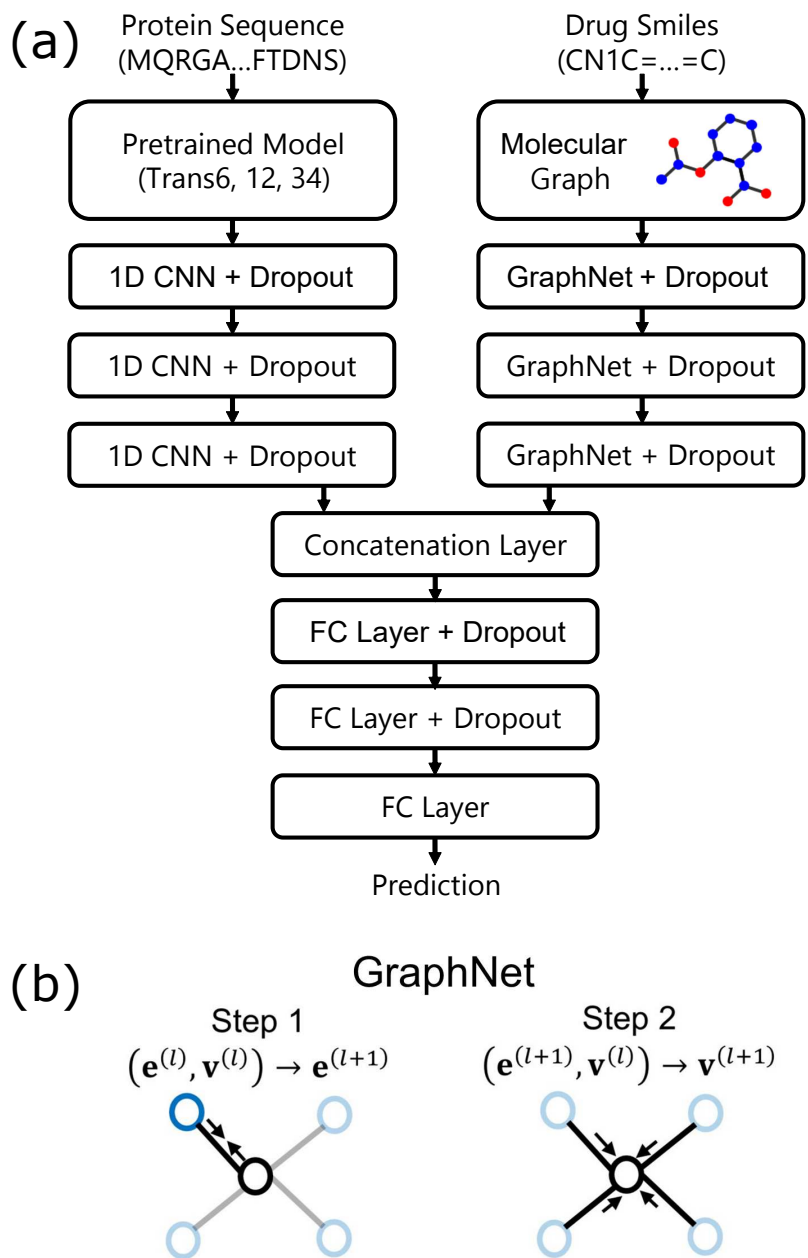}
	\centering
	\caption{An overview of schematic of proposed neural network architecture. (a) The protein and drug representations are obtained by passing the pretrained transformer model and GraphNet layers, respectively. The protein and drug representation vectors are concatenated and fed into a classifier consisting of fully-connected layers. (b) Mechanism of the message passing in GraphNet. The GraphNet performs message passing on the molecular graph, recursively updating graph edges $\mathbf{e}^{(l)}$ and nodes $\mathbf{v}^{(l)}$.}
	\label{fig:schematic}
\end{figure}

In this study, the DPI is defined as a binary label, which represents the presence of an interaction. Figure \ref{fig:schematic} (a) shows schematic of proposed model. The input data is a pair of strings consisting of a protein sequence and drug SMILES strings. The input data passes embedding layers to be encoded as a pair of representation vectors. These protein and drug representation vectors are then concatenated and passed through fully-connected layers, resulting in a prediction for the existence of an interaction. In each cycle of training, this prediction is compared with the ground truth, and model parameters are tuned to decrease the difference between the two using the backpropagation algorithm. To implement BNNs, we apply dropout layers in every layer except the pretrained layer, the concatenation layer, and the final fully-connected layer. Detailed descriptions of the model are given below.
 
 \subsubsection*{Feature extraction of protein}
 A protein sequence is represented as a list of amino acids provided in the raw training data. Note that we do not use a set of gene ontology annotations that provides high-level information on protein functions. To extract protein-level embeddings, we use the pretrained models from Rives \textit{et al.}\cite{Rives2019}, which were trained with 250 million protein sequences in an unsupervised manner. Rives \textit{et al.} used an attention-based transformer architecture\cite{Vaswani2017}, and found that their model outperforms other recurrent network-based methods for predicting protein functionality. We select three models, Trans6, Trans12, and Trans34, which are pretrained with 6, 12, and 34 transformer layers, respectively.
 
 For each protein sequence of length $L_{\textrm{p}}$, the pretrained models outputs an embedding matrix $\mathbf{X}_{\textrm{p}}\in \mathbb{R}^{L \times d}$, where $d = 768$ for Trans6, Trans12 and $d = 1,280$ for Trans34 model. From amino-acid level feature $\mathbf{X}_{\textrm{p}}$, we obtain the protein level feature $\mathbf{x}^{(0)}_{\textrm{p}} \in \mathbb{R}^{d}$ by averaging over the $L$ amino acids features. 
 
 With the protein-level embedding $\mathbf{x}^{(0)}_{\textrm{p}}$, we use three 1-dimensional convolutional neural networks (1D-CNN) to smooth patterns in protein features. Note that the 1D-CNN gives slightly better performance than the fully-connected layers. 
 
 \subsubsection*{Feature extraction of drug} 
 The raw training data of drugs is in the SMILES (Simplified Molecular Input Line Entry System) format\cite{Weininger1988}. For each input SMILES string, we construct a corresponding molecular graph that contains connectivity and structure information of the compound.   
 
 In the molecular graph, atoms and bonds are represented with vectors with structural features that characterize the surrounding chemical environment. The details of the attributes are shown in Supplmentary Table 1, which follows the feature design from DeepChem\cite{Wu2018}. The graph construction and corresponding feature extraction processes are conducted using RDKit\cite{Landrum2006} - an open-source chemical informatics software. Initial encoding of the $i$-th atom and bond between the $i$- and $j$-th atoms are denoted as vectors, $\mathbf{v}^{(0)}_{i}$ and $\mathbf{e}^{(0)}_{ij}$, respectively. These atom and bond features are updated by a message passing-based graph network during model inference.
 
 The message passing framework of graph data has been used broadly to predict the properties of crystal\cite{Xie2018}, organic molecules\cite{Ryu2019}, ice\cite{Kim2020}, and glasses\cite{Bapst2020}.  To extract the drug molecule features, we use the graph interaction network (GraphNet) model\cite{Battaglia2016}. Figure \ref{fig:schematic} (b) shows the schematic of the GraphNet mechanism. First proposed by Battaglia \textit{et al.}\cite{Battaglia2016} to infer interaction between objects, the GraphNet exchanges information between graph edges and nodes and recursively updates them. 
 
 The GraphNet first updates an edge between $i$- and $j$-th node as,
\begin{equation}
 \mathbf{e}^{(l+1)}_{i} = \textrm{ReLU} \left[ \left( \mathbf{e}^{(l)}_{ij} \oplus \mathbf{v}^{(l)}_{i} \oplus \mathbf{v}^{(l)}_{j} \right) \mathbf{W}^{(l)}_{\textrm{e}} + \mathbf{b}^{(l)}_{\textrm{e}} \right],
\end{equation}
 where $\oplus$ is the concatenation operator, $\mathbf{W}^{(l)}_{\textrm{e}}$ is the weight matrix of the edge update, and $\mathbf{b}^{(l)}_{\textrm{e}}$ is the bias. Then update of the $i$-th node is carried out with the features of the node and on the sum of its linked edge features as,
\begin{equation}
 \mathbf{v}^{(l+1)}_{i} = \textrm{ReLU} \left[ \left( \mathbf{v}^{(l)}_{i} \oplus \sum_{j \in N(i)} \mathbf{e}^{(l+1)}_{ij}   \right) \mathbf{W}^{(l)}_{\textrm{v}} + \mathbf{b}_{\textrm{v}}^{(l)} \right],
\end{equation}
 where $\mathbf{W}^{(l)}_{\textrm{v}}$ is the weight matrix of node update, and $\mathbf{b}_{\textrm{v}}^{(l)}$ is the bias. After the update of node states is finalized, we obtain a graph feature (molecular feature) by gathering all the node and edge states. We choose most typical readout function, which is an average of all atom states processed by,
 \begin{equation}
 \mathbf{x}_{\textrm{d}} = \frac{1}{N} \sum_{i} \left( \mathbf{v}_{i} \oplus \mathbf{e}_{i} \right),
 \end{equation}
 where $N$ is the number of nodes in the molecular graph.
 
 \subsubsection*{Classifier}
 We prepare the drug-protein feature vector $\mathbf{x}$ by concatenating $\mathbf{x}_{\textrm{p}}$ and $\mathbf{x}_{\textrm{d}}$,
 \begin{equation}
 \mathbf{x} = \mathbf{x}_{\textrm{p}} \oplus \mathbf{x}_{\textrm{d}}.
 \end{equation}
 In the classifier block, the feature vector $\mathbf{x}$ passes fully connected (FC) layers with ReLU activation to output final prediction value. The dimension of the last layer is 2, corresponding to the one-hot encoding of the binary classification labels.

\subsubsection*{Bayesian neural network}
For a given training set $\left\{ \mathbf{X}, \mathbf{Y} \right\}$, let $p\left( \mathbf{Y} | \mathbf{X}, \mathbf{w} \right)$ and $p\left( \mathbf{w} \right)$ be model likelihood and prior distribution for a vector of model parameters $\mathbf{w}  = \{\mathbf{W}_{1}, ..., \mathbf{W}_{k}\}$, where $k$ is a number of layers. In a Bayesian framework, model parameters are considered as random variables and the output is defined as
\begin{equation}
\label{eqn:predictive_distribution}
p\left( \mathbf{y}^{*} | \mathbf{x}^{*}, \mathbf{X}, \mathbf{Y} \right) = \int_{\Omega} p\left( \mathbf{y}^{*}|\mathbf{x}^{*}, \mathbf{w}\right) p\left(\mathbf{w} | \mathbf{X}, \mathbf{Y} \right) \rm{d}\mathbf{w}
\end{equation}
for a new input $\mathbf{x}^{*}$ and a new output $\mathbf{y}^{*}$. 

The direct computation of Eq. \eqref{eqn:predictive_distribution} in neural network is often infeasible because the heavy computational cost is required to train ensemble of weights. Here, we use a variational inference that approximates the posterior distribution with a distribution $p\left(\mathbf{w} | \mathbf{X}, \mathbf{Y} \right) \sim q_{\theta}\left( \mathbf{w} \right)$ parameterized by a small-dimensional variational parameter $\theta$. 
 The quality of variational distribution $q_{\theta}\left(\mathbf{w}\right)$ is crucial to the implementation of BNN. The recently proposed Monte-Carlo dropout (MC-dropout) approach attaches dropout layer to every neural network layers to approximate the posterior distribution with a product of Bernoulli distributions\cite{Gal2016}. The MC-dropout method is practical because it does not need ensemble of the models to obtain the variational posterior distribution. Also, the expectation and variance of output can be easily obtined with the collection of outputs sampled by repeated inference of new input $\textbf{x}^{*}$ while the dropout layers are turned on. Thus, we adopt MC-dropout in this work.

A variational inference approximating a posterior a variational distribution $q_{\theta}\left(\mathbf{w}\right)$ provides a variational predictive distribution of a new output $\mathbf{y}^{*}$ given a new input $\mathbf{x}^{*}$ as
\begin{equation}
\label{eqn:prediction}
q^{*}_{\theta}\left(\mathbf{y}^{*}|\mathbf{x}^{*}\right) = \int_{\Omega} q_{\theta}\left(\mathbf{w}\right)p\left(\mathbf{y}^{*}|\hat{\mathbf{y}}(\mathbf{w})_{t}^{*}\right) \textrm{d}\mathbf{w},
\end{equation}
where $\hat{\mathbf{y}}(\mathbf{w})_{t}^{*}$ is a output of input $\mathbf{x}_{t}^{*}$ with a given $\mathbf{w}$. In BNN, the integration in Eq. \eqref{eqn:prediction} is replaced with a predictive mean of $T$ times of MC sampling, which is estimated by
\begin{equation}
\label{eqn:mean}
\hat{E}\left[\mathbf{y}^{*} | \mathbf{x}^{*}\right] = \frac{1}{T} \sum_{t = 1}^{T} \hat{\mathbf{y}}_{t}^{*}.
\end{equation}

In estimating its predictive variance, we decompose the source of uncertainty into aleatoric and epistemic, which was first suggested by Kendall and Gal\cite{Kendall2017} and optimized for classification tasks by Kwon \textit{et al.}\cite{Kwon2020}. The aleatoric uncertainty originates from the inherent noise of data points, and the epistemic uncertainty arises due to model prediction variability. Here we use the method suggested by Kwon \textit{et al.}\cite{Kwon2020}, which does not involve extra variance parameters at the last layer.

The predictive variance is estimated by
\begin{equation}
\label{eqn:variance}
\hat{\textrm{Var}} \left[\mathbf{y}^{*} | \mathbf{x}^{*}\right] = \underbrace{ \frac{1}{T} \sum_{t = 1}^{T} \left(\hat{\mathbf{y}}^{*}_{t} - \bar{\mathbf{y}}\right) \left(\hat{\mathbf{y}}^{*}_{t} - \bar{\mathbf{y}}\right)^{T} }_\text{epistemic} + \underbrace{ \frac{1}{T}\sum_{t = 1}^{T} \left( \textrm{diag}\left(\hat{\mathbf{y}}_{t}^{*}\right) - \left(\hat{\mathbf{y}}_{t}^{*}\right)\left(\hat{\mathbf{y}}_{t}^{*}\right)^{T} \right) }_\text{aleatoric}.
\end{equation}

\subsection*{Implementation and Evaluation Strategy}
 We implement the proposed model with Pytorch 1.5.1\cite{Paszke2019}. The training process takes at most 200 epochs on all the datasets using the Adam optimizer\cite{Kingma2014} with a learning rate of 0.001 and a batch size of 32. The hidden layer dimensions of GraphNet in the drug feature extractor and MLP in the classifier are 256 and 512, respectively. The number of layers of both the protein and drug feature extractor is set to 3. The coefficient of L2 regularization is 0.001. These hyperparameters are searched in a wide range.

The training objective is to minimize the loss function $\mathcal{L}$, given by the sum of the cross-entropy loss and the regularization as follows
\begin{equation}
\mathcal{L}\left( \mathbf{w} \right) = -\sum_{i = 1}^{N} y_{i} \left[ \log \hat{y}_{i} + \left(1-y_{i}\right) \log \left(1 - \hat{y}_{i} \right) \right] + \frac{\lambda}{2}\| \mathbf{w} \|^{2}_{2},
\end{equation}
where $\mathbf{w}$ is the set of model parameters, $N$ is the number of interaction labels, and $\lambda$ is an L2 regularization hyperparameter.

 To implement MC-dropout sampling, we turn on dropout layers in estimating test dataset with $T = 30$ samplings. The mean performance and the decomposed uncertainties of the output are calculated with Eq. \eqref{eqn:mean} and Eq. \eqref{eqn:variance}, respectively.

 The main performance metric was chosen to be the area under the receiver operating curve (ROC-AUC). We also report some additional performance metrics - accuracy for the BindingDB dataset, and precision and recall for the Human and \textit{C. elegans} dataset.

\section*{Results}
To train DPI datasets, we prepare six models, Trans6, Trans12, Trans34, Trans6+Drop, Trans12+Drop, and Trans34+Drop. The latter three models use the pretrained protein model and implement the BNN architecture with MC-Dropout (Fig. \ref{fig:schematic} (a)), while the former three models only use the pretrained model. The numbers 6, 12, and 34 correspond to the number of transformer layers in the pretrained model.

 \subsection*{Performance of proposed model}
 With the BindingDB dataset, we compare our model against three baselines: Tiresias, DBN, and E2E. Tiresias uses similarity measures of drug and protein pairs\cite{Fokoue2016}. DBN uses stacked restricted Boltzmann machines with the inputs as extended connectivity fingerprints\cite{Wen2017}. E2E uses graph convolutional networks and LSTM to process drug-protein pair information with Gene Ontology annotations\cite{Gao2018}.
 
 Following suggestions from previous works, we further split the test dataset into four sub-test sets that the model can be learned and applied to predict the label between a drug and protein target. The binary interaction test data is divided by ``seen'' and ``unseen'' whether the protein and drug are observed in the training dataset. 
 
\begin{figure}[h]
	\includegraphics[width=0.9\textwidth]{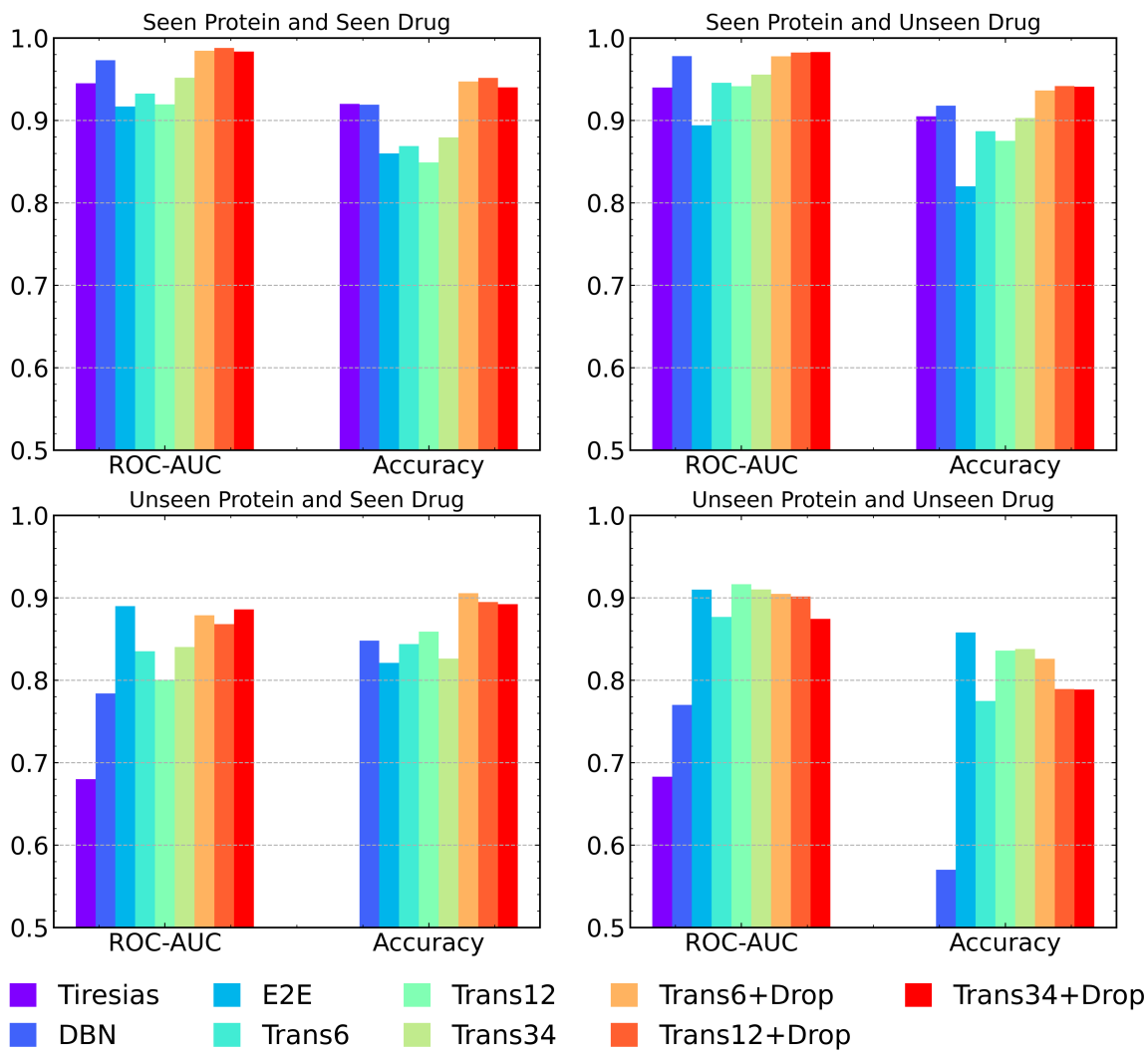}
	\centering
	\caption{Performance comparison of proposed models, similarity-based approach (Tiresias), stacked restricted Boltzmann layers (DBN), and graph convolutional networks - long short-term memory-based approach (E2E). For each model, two metrics are reported: area under receiver operating characteristic curve (ROC-AUC) and accuracy. The binary interaction test data is divided by ``seen'' and ``unseen'' whether the protein and drug are observed in the training dataset. The accuracy score of Tiresias is not seen in the bottom graphs because they are lower than the lower bound of $y$-axis.}
	\label{fig:score_bind}
\end{figure}

Figure \ref{fig:score_bind} shows that the proposed method consistently performs well on all four sub-test sets. The models with pretraining and MC-dropout give a high performance consistently in four categories. The sub-test dataset with unseen protein is difficult to classify, while only the E2E model shows comparable performance with our proposed model. Tiresias and DBN perform well on seen proteins and outperform E2E but have much worse performance on unseen proteins because these models are overfitted. The score of the unseen protein dataset is consistently lower than that of the unseen drug dataset. It implies that the extraction of generalized protein embedding with a long sequence plays an important role in DPI classification. If we measure scores with aggregating four test sub-datasets, the ROC-AUC of Trans6+Drop, which achieves the best score in the proposed model, is 0.943 while that of E2E is 0.913 and DBN is 0.817.
 
 Also, we compare the proposed method with the previous DPI approaches on the Human and the \textit{C. elegans} dataset. We compare it with k-nearest neighbor (k-NN), random forest (RF), L2-logistic (L2), support vector machine (SVM), and graph neural network (GNN) models. Note that the GNN model uses $n$-grams to embed protein sequence.
 
 As shown in Tables \ref{tbl:score_human}, the our best performing model achieves the highest AUC, precision, and recall scores among the neural network-based method. In the human dataset, SVM shows better performance in the Precision score, but the proposed model outperforms the other metrics. In the \textit{C. elegans} dataset, Trans6+Drop shows the best performance all metrics, except the recall score of the balanced dataset that Trans34+Drop is the best. 
 
 Our results show that models with both transfer learning and BNN (Trans6+Drop, Trans12+Drop, Trans34+Drop) outperform other baseline models when evaluated with three public DPI datasets. We note that only the pretrained protein sequence can train models (Trans6, Trans12, Trans34) competitive with the baselines, but additional Bayesian frameworks further increase performance. It suggests that the role of BNN, training robust model, is the key figure of performance enhancement as well as the expression capacity obtained from the pretrained model. 
 
 An additional point to mention is that the most complex model, Trans34+Drop, does not always give the best results. This is in agreement with the literature, where it was found that the prediction accuracy is not strictly proportional to the sequence model complexity\cite{Rives2019}. Therefore, when using transfer learning, we recommend preparing several different pretrained models and comparing their results before making the final choice.

 \subsection*{Robustness of proposed model}
 In this section, we test the robustness of the Bayesian models by varying the quality of the protein data. The robustness is estimated by tracking the degradation of the model performance as more and more external noise is added to the dataset. The type of noise for the experiment is chosen to be the Gaussian noise $\mathcal{N}\left(0, \sigma^{2}\right)$, where $0$ is the mean and $\sigma$ is the standard deviation of the distribution.
 
\begin{figure}[h]
	\includegraphics[width=0.9\textwidth]{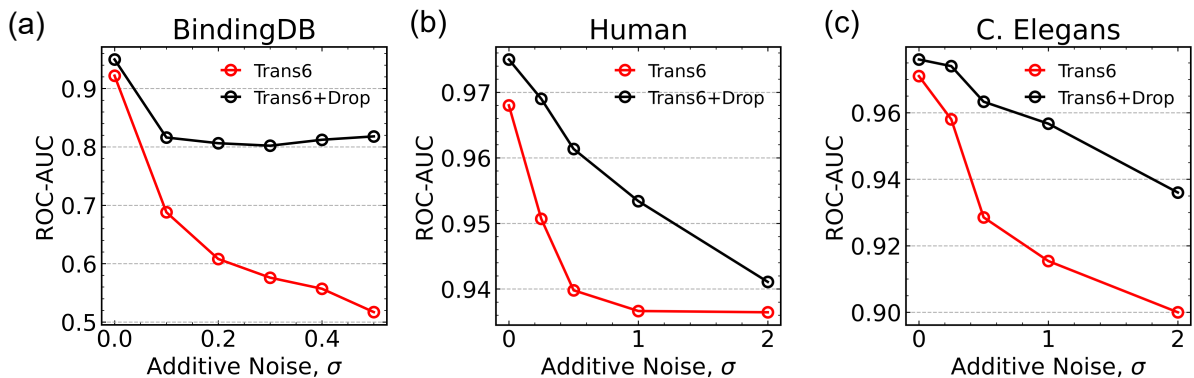}
	\centering
	\caption{ROC-AUC scores on the test set as a function of standard deviation of the additive noise on (a) BindingDB, (b) Human, and (c) \textit{C. Elegans} dataset. The additive noise is sampled from the Gaussian distribution $\mathcal{N}\left(0, \sigma^{2}\right)$.}
	\label{fig:noise}
\end{figure}

 Figure \ref{fig:noise} shows the ROC-AUC scores of the two models Trans6 and Trans6+Drop applied to three DPI datasets as a function of the noise level $\sigma$. As the noise level increases, the ROC-AUC of Trans6+Drop is more robust to the additive noise than that of Trans6. In the BindingDB dataset, the ROC-AUC score of Bayesian Trans6+Drop does not fall under 0.8 when noise standard deviation increases until 0.5, where Trans6 loses its predictability. For Human and \textit{C. Elegans} datasets, the models maintain relatively good performance regardless of the additive noise, where the Bayesian model outperforms the other. It indicates that the BNN architecture trains more robust model and it attributes the overall enhanced performance of our proposed model.

 \subsection*{Quality of estimated uncertainties}
 We first test whether the uncertainties obtained from the proposed BNN model are correctly estimated. This is accomplished by reducing the training set sizes and observing the resulting uncertainty changes. When dataset size is decreased, aleatoric uncertainty, which is related to the inherent noise of the data, should stay constant, whereas the model error-related epistemic noise should increase to a lack of sufficient training data.
 
 Table \ref{tbl:uncertainty} shows the uncertainties obtained from the reduced training set sizes (1, 1/2, 1/4) and the entire test set. The uncertainties are obtained via Eq. \eqref{eqn:variance}. It shows that the epistemic uncertainty increases as the training size gets larger, while the aleatoric uncertainty remains relatively constant. It indicates that our proposed model reliably estimate uncertainties.    

\begin{figure}[h]
	\includegraphics[width=0.9\textwidth]{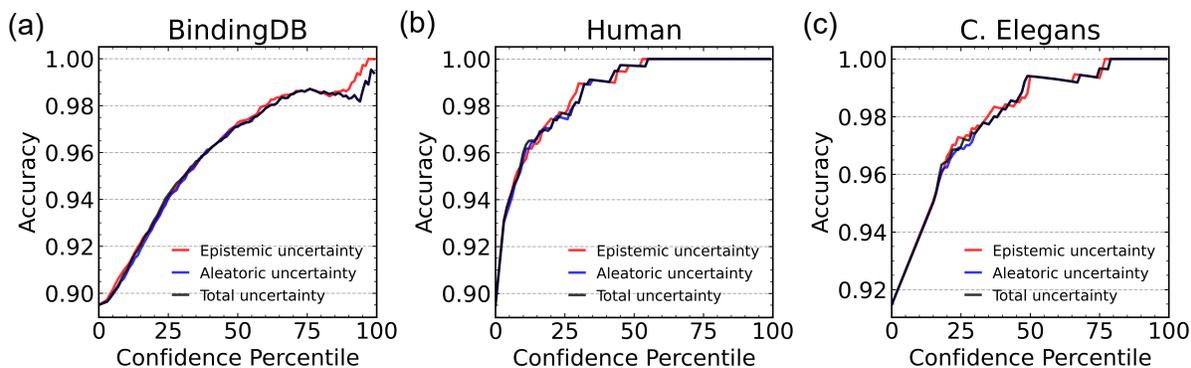}
	\centering
	\caption{Model accuracies on the test set as a function of confidence percentile of (a) BindingDB, (b) Human, and (c) \textit{C. Elegans} dataset. The confidence is estimated based on the epistemic uncertainty (red line), aleatoric uncertainty (blue line), and sum of the two (black line).}
	\label{fig:uncertainty}
\end{figure}

 Because the model successfully estimates uncertainties, we can plot confidence-accuracy graphs, as shown in Fig. \ref{fig:uncertainty}. We use three uncertainties, an epistemic uncertainty, aleatoric uncertainty, and the sum of two. Here the confidence percentile means that we only consider the top $n$ percent of data points in the test set ranked by the confidence, which is defined by the inverse of uncertainty. The plots show how the test set accuracy varies as a function of the confidence percentile. In every dataset, the accuracy is an increasing function of model confidence. Thus the data points with low confidence can be interpreted as the outlier and can be screened in DPI datasets in drug development applications. For example, if we delete 50 $\%$ of the lowest confident points of the Human dataset, we can achieve nearly 100 $\%$ accuracy. Note that there is no consistent trend regarding which uncertainty is more important, and two uncertainties should be treated equally to achieve an accurate estimation.

\section*{Conclusion}
 In this study, we present a novel Bayesian deep learning framework with a pretrained protein sequence model to predict drug-protein interactions. Experiments on three public datasets demonstrate that our proposed model consistently outputs increased prediction accuracies. Our estimation of model performance shows that Bayesian neural networks are highly robust to additive noise, which explains the superior performances of the proposed model. Furthermore, from the prediction uncertainty our model outputs, one can evaluate the confidence level of a dataset, which can then be used to screen the dataset for unreliable data points.

\section*{Code availability}
The code is available in https://github.com/QHwan/PretrainDPI.


\section*{Acknowledgements}
This work was supported by the National Research Foundation of Korea (NRF) grant funded by the Korea Government (MSIP) (No. 2016R1A3B1908660).

\section*{Author contributions statement}
All authors contributed to construct concept and initialize the project. Q.K and W.J made the program. All authors participated in the discussion of the results. Q.K and W.J wrote the manuscript. All authors reviewed the manuscript.  

\section*{Competing interests}
The authors declare no competing interests.

\begin{table}[h]
	\caption{ROC-AUC, Precision, and Recall scores of human and \textit{C. elegans} dataset with proposed models, k-nearest neighbor (k-NN), random forest (RF), L2 logistic (L2), support vector machine (SVM), and graph neural network (GNN) proposed by Tsubaki \textit{et al.}\cite{Tsubaki2019}. The best score of proposed models is emphasized in bold. The italicized scores correspond to the best scores for the baseline models.}
	\label{tbl:score_human}
	\begin{tabular}{|c|c c c|c c c|}
		\multicolumn{7}{c}{Human} \\
		\hline
		 & \multicolumn{3}{|c|}{Balanced Dataset (1 : 1)} & \multicolumn{3}{|c|}{Unbalanced Dataset (1 : 3)}  \\
		\hline
		Methods  & ROC-AUC & Precision & Recall & ROC-AUC & Precision & Recall  \\
		\hline
		KNN & 0.860 & 0.798 & 0.927 & 0.904 & 0.716 & 0.882 \\
		RF    & 0.940 & 0.861 & 0.897 & \textit{0.954} & 0.847 & 0.824 \\
		L2    & 0.911 & 0.891 & 0.913 & 0.920 & 0.837 & 0.773 \\
		SVM & 0.910 & \textit{0.966} & \textit{0.950} & 0.942 & \textit{0.969} & 0.883 \\
		GNN & \textit{0.970} & 0.923 & 0.918 & 0.950 & 0.949 & \textit{0.913}\\
		\hline
		Trans6 & 0.968 & 0.902 & 0.901 & 0.971 & 0.915 & 0.910 \\
		Trans12 & 0.960 & 0.881 & \textbf{0.949} & 0.969 & \textbf{0.958} & 0.863 \\
		Trans34 & 0.973 & 0.914 & 0.925 & 0.971 & 0.930 & 0.863 \\
		Trans6+Drop & \textbf{0.975} & 0.932 & 0.922 & \textbf{0.976} & 0.939 & 0.902 \\
		Trans12+Drop & 0.971 & 0.914 & 0.924 & 0.963 & 0.932 & 0.902 \\
		Trans34+Drop & \textbf{0.975} & \textbf{0.945} & 0.935 & 0.970 & 0.925 & \textbf{0.923} \\
		\hline
		\multicolumn{7}{c}{}\\
		\multicolumn{7}{c}{\textit{C. elegans}} \\
		\hline		
		& \multicolumn{3}{|c|}{Balanced Dataset (1 : 1)} & \multicolumn{3}{|c|}{Unbalanced Dataset (1 : 3)}  \\
		\hline
		Methods  & ROC-AUC & Precision & Recall & ROC-AUC & Precision & Recall  \\
		\hline
		KNN & 0.858 & 0.801 & 0.827 & 0.892 & 0.787 & 0.743 \\
		RF    & 0.902 & 0.821 & 0.844 & 0.926 & 0.836 & 0.705 \\
		L2    & 0.892 & 0.890 & 0.877 & 0.896 & 0.875 & 0.681 \\
		SVM & 0.894 & 0.785 & 0.818 & 0.901 & 0.837 & 0.576 \\
		GNN & \textit{0.978} & \textit{0.938} & \textit{0.929} & \textit{0.971} & \textit{0.916} & \textit{0.921} \\
		\hline
		Trans6 & 0.981 & 0.937 & 0.949 & 0.977 & 0.871 & 0.917 \\
		Trans12 & 0.975 & 0.949 & 0.910 & 0.967 & 0.876 & 0.861 \\
		Trans34 & 0.973 & 0.914 & 0.925 & 0.969 & 0.900 & 0.915 \\
		Trans6+Drop & \textbf{0.986} & \textbf{0.955} & 0.933 & \textbf{0.983} & \textbf{0.923} & \textbf{0.944} \\
		Trans12+Drop & 0.980 & 0.946 & 0.928 & 0.981 & 0.890 & 0.940 \\
		Trans34+Drop & 0.981 & 0.946 & \textbf{0.940} & 0.980 & 0.914 & 0.937 \\
		\hline

	\end{tabular}
	\centering
\end{table}

\begin{table}[h]
	\caption{Epistemic and aleatoric uncertainties for a range of different training dataset sizes (1, 1/2, 1/4 of the original training dataset size)
 The results show that the aleatoric uncertainty remains approximately constant, whereas the epistemic uncertainty increases when the training size decreases.}
	\label{tbl:uncertainty}
	\begin{tabular}{lll}
		\hline
		Dataset & Epistemic & Aleatoric  \\
		\hline
		BindingDB / 4 & 0.018 & 0.036 \\
		BindingDB / 2 & 0.013 & 0.037 \\
		BindingDB & 0.011 & 0.037 \\ 
		\hline
		Human / 4 & 0.0128 & 0.020 \\
		Human / 2 & 0.0096 & 0.018 \\
		Human & 0.0082 & 0.019 \\
		\hline
		\textit{C. elegans} / 4 & 0.0137 & 0.0155 \\
		\textit{C. elegans} / 2 & 0.0098 & 0.0153 \\
		\textit{C. elegans} & 0.0053 & 0.0143 \\
		\hline
	\end{tabular}
	\centering
\end{table}

\end{document}